\documentclass[acmtog]{acmart}
\acmSubmissionID{334}

\usepackage{booktabs} 

\usepackage{bbm}
\usepackage{amsfonts}
\usepackage{amsthm}
\usepackage{multirow}
\usepackage{enumitem}
\usepackage{framed}
\usepackage{mathtools} 
\usepackage{breqn}
\usepackage{footmisc}
\usepackage{makecell}
\usepackage{footnote}
\usepackage[ruled]{algorithm2e}

\SetAlFnt{\small}
\SetAlCapFnt{\small}
\SetAlCapNameFnt{\small}
\SetAlCapHSkip{0pt}

\definecolor{myblue}{rgb}{0, 0.44, 0.74}
\newcommand{\ie}{\emph{i.e}.}
\newcommand{\eg}{\emph{e.g}.}

\setcopyright{acmcopyright}\acmJournal{TOG}
\acmYear{2022}\acmVolume{41}\acmNumber{4}\acmArticle{162}\acmMonth{7} \acmDOI{10.1145/3528223.3530104}

\begin{document}
\title{Text2Human: Text-Driven Controllable Human Image Generation}

\author{Yuming Jiang}
\orcid{0000-0001-7653-4015}
\affiliation{
 \institution{S-Lab, Nanyang Technological University}
 \country{Singapore}}
\email{yuming002@e.ntu.edu.sg}
\author{Shuai Yang}
\orcid{0000-0002-5576-8629}
\affiliation{
 \institution{S-Lab, Nanyang Technological University}
 \country{Singapore}
}
\email{shuai.yang@ntu.edu.sg}
\author{Haonan Qiu}
\orcid{0000-0002-3878-1418}
\affiliation{
\institution{S-Lab, Nanyang Technological University}
\country{Singapore}}
\email{haonan002@e.ntu.edu.sg}
\author{Wayne Wu}
\orcid{0000-0002-1364-8151}
\affiliation{
 \institution{SenseTime Research}
 \country{China}
}
\email{wuwenyan0503@gmail.com}
\author{Chen Change Loy}
\orcid{0000-0001-5345-1591}
\affiliation{
 \institution{S-Lab, Nanyang Technological University}
 \country{Singapore}}
\email{ccloy@ntu.edu.sg}
\author{Ziwei Liu}
\authornote{Corresponding Author}
\orcid{0000-0002-4220-5958}
\affiliation{
 \institution{S-Lab, Nanyang Technological University}
 \country{Singapore}
}
\email{ziwei.liu@ntu.edu.sg}

\begin{abstract}
Generating high-quality and diverse human images is an important yet challenging task in vision and graphics. 
However, existing generative models often fall short under the high diversity of clothing shapes and textures. 
Furthermore, the generation process is even desired to be intuitively controllable for layman users. 
In this work, we present a text-driven controllable framework, Text2Human, for a high-quality and diverse human generation. We synthesize full-body human images starting from a given human pose with two dedicated steps. 1) With some texts describing the shapes of clothes, the given human pose is first translated to a human parsing map. 
2) The final human image is then generated by providing the system with more attributes about the textures of clothes. 
Specifically, to model the diversity of clothing textures, we build a hierarchical texture-aware codebook that stores multi-scale neural representations for each type of texture.
The codebook at the coarse level includes the structural representations of textures, while the codebook at the fine level focuses on the details of textures.
To make use of the learned hierarchical codebook to synthesize desired images, a diffusion-based transformer sampler with mixture of experts is firstly employed to sample indices from the coarsest level of the codebook, which then is used to predict the indices of the codebook at finer levels.
The predicted indices at different levels are translated to human images by the decoder learned accompanied with hierarchical codebooks.
The use of mixture-of-experts allows for the generated image conditioned on the fine-grained text input.
The prediction for finer level indices refines the quality of clothing textures.
Extensive quantitative and qualitative evaluations demonstrate that our proposed Text2Human framework can generate more diverse and realistic human images compared to state-of-the-art methods. Our project page is {\textcolor{myblue}{\url{https://yumingj.github.io/projects/Text2Human.html}}}. Code and pretrained models are available at {\textcolor{myblue}{\url{https://github.com/yumingj/Text2Human}}}.
\end{abstract}

%



\begin{teaserfigure}
  \includegraphics[width=\textwidth]{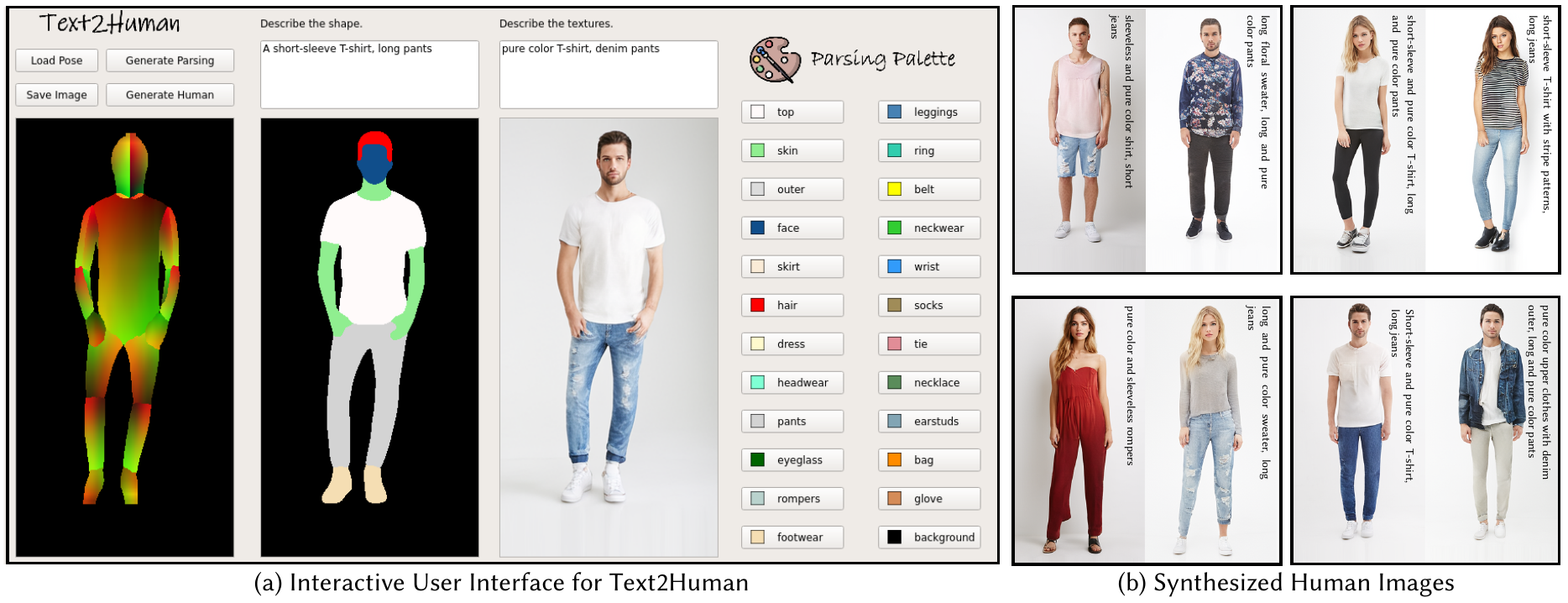}
  \caption{\textbf{Text-Driven Controllable Human Image Generation.} (a) User Interface. (b) Synthesized Human images driven by texts at the right side.}
  \label{fig:teaser}
\end{teaserfigure}

\maketitle

\section{Introduction}

Recent years have witnessed the rapid progress of image generation since the emergence of Generative Adversarial Networks (GANs) \cite{goodfellow2014generative}.
Nowadays, we can easily generate diverse faces of high fidelity using a pretrained StyleGAN \cite{karras2020analyzing}, which further supports several downstream tasks, such as facial attribute editing \cite{abdal2021styleflow, patashnik2021styleclip,jiang2021talk}
and face stylization \cite{song2021agilegan, pinkney2020resolution,yang2022Pastiche}.

Human full-body images, another type of human-related media, are more diverse, richer, and fine-grained in content. Furthermore, human image generation \cite{fu2022styleganhuman,Fruehstueck2022InsetGAN,grigorev2021stylepeople} has wide applications, including human pose transfer \cite{albahar2021pose, sarkar2021style}, virtual try-on \cite{lewis2021tryongan, cui2021dressing}, and animations \cite{yoon2021pose,chan2019everybody,hong2022avatarclip}. 
From the perspective of applications and interactions, apart from generating high-fidelity human images, it is even desirable to intuitively control the synthesized human images for layman users. For example, they may want to generate a person wearing a floral T-shirt and jeans without expert software knowledge.
Human image generation with explicit textual controls makes it possible for users to create 2D avatars more easily.
    
Despite the great potential, controllable human body image generation with high fidelity and diversity is less explored due to the following challenges:
\textbf{1)} Compared to faces, human body images are more complex with multiple factors, including the diversity of human poses, the complicated silhouettes of clothing, and sundry textures of clothing; 
\textbf{2)} Existing human body image generation methods \cite{sarkar2021humangan, yildirim2019generating, weng2020misc} fail to generate diverse styles of clothes since they tend to generate clothes with simple patterns like pure color, let alone fine-grained controls on the textures of clothes in the generated images.
\textbf{3)} The generation of clothes with textual controls relies on additional fine-grained annotations. However, currently, there is a lack of human image generation datasets containing fine-grained labels on clothes shapes and textures \cite{liu2016deepfashion,liu2016fashion,cai2022humman}.
To bridge the gap, in this work, we propose the Text2Human framework for the text-driven controllable human image generation. As shown in Fig.~\ref{fig:teaser}, given a human pose, users can specify the clothes shapes and textures using solely natural language descriptions. Human images are then synthesized in accordance with the textual requests.

Due to the complexity of human body images, it is challenging to handle all involving factors in a single generative model. We decompose the human generation task into two stages.
Stage I generates a human parsing mask with diverse clothes shapes based on the given human pose and user-specified texts describing the clothes shapes.
Then Stage II enriches the human parsing mask with diverse textures of clothes based on texts describing the clothes textures.

Considering the high diversity of clothes textures, we introduce the concept of codebook, which is widely used in VQVAE-based methods \cite{oord2017neural, esser2021taming}, into our framework.
The codebook learns discrete neural representations of images.
To adaptively characterize textures, we propose a hierarchical VQVAE with texture-aware codebook designs. Specifically, the codebooks are constructed in multiple scales. The codebook in the coarser scale contains more structural information about textures of clothes, while the codebook in finer scales includes more detailed textures.
Due to the different natures of different textures, we also build codebooks separately for each texture.

In order to conditionally generate human images consistent with the texts describing the textures, we need a sampler to select appropriate texture representations (\ie, codebook indices) from the codebook, and then re-arrange them in a reasonable order in the spatial domain. In this manner, with rich texture representations stored in codebooks, the human generation task is formulated as to sample an intermediate feature map from the learned codebooks.
We adopt the diffusion model based transformer \cite{bond2021unleashing, gu2021vector, esser2021imagebart} as the sampler. 
With the texture-aware codebook design, we incorporate mixture-of-experts \cite{shazeer2017outrageously} into the sampler. The sampler has multiple index prediction expert heads to predict indices for different textures. 

With the hierarchical codebooks, we need to sample intermediate feature maps from the coarse level to the fine level, \ie, sampling indices for both the coarse-level and fine-level codebook is required for the image synthesis.
Thanks to the implicit relationship between codebooks at different levels learned by our proposed hierarchical VQVAE, the indices of codebook at the coarse level can provide hints for the sampling of the fine level features.
A similar idea is also adopted VQVAE2 \cite{razavi2019generating}. However, in VQVAE2, the pixel-wise sampling by auto-regressive models is time-consuming. 
By comparison, we propose a feed-forward codebook index prediction network, which predicts the desired fine-level codebook indices directly from the coarse-level features. The proposed index prediction network speeds up the sampling process and ensures the generation quality.

To facilitate the controllable human generation, we construct a large-scale full-body human image dataset dubbed DeepFashion-MultiModal dataset, 
which contains rich clothes shape and texture annotations, human parsing masks with diverse fashion attribute classes, and human poses. Both the textual attribute annotations and human parsing masks are manually labeled. The human poses are extracted using \cite{guler2018densepose}. 
All images are collected from the high-resolution version of DeepFashion dataset. 
These images are further cleaned and selected to ensure they are full-body and of good quality 
\footnote{The dataset is available at {\textcolor{myblue}{\url{https://github.com/yumingj/DeepFashion-MultiModal}}}.}.

\begin{figure*}
   \begin{center}
      \includegraphics[width=1.0\linewidth]{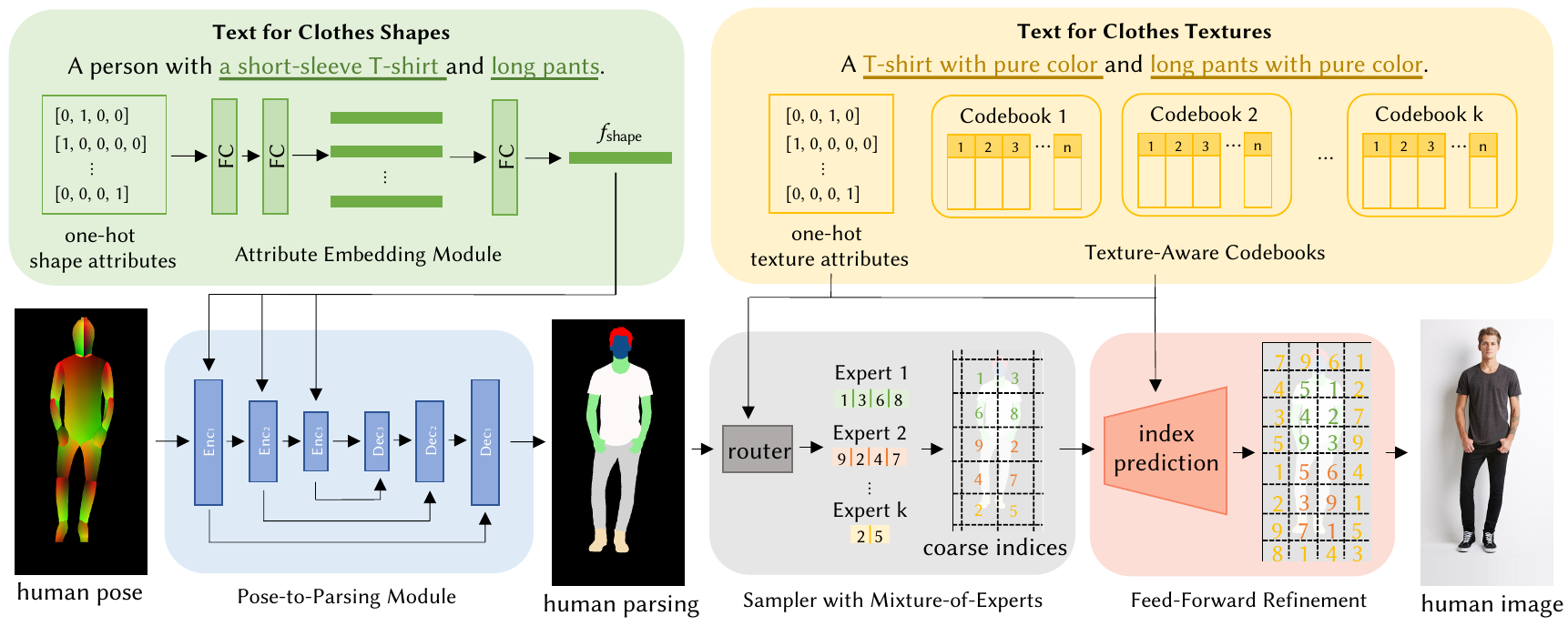}
   \end{center}
   \caption{\textbf{Overview of Text2Human.} We decompose the human generation into two stages. Stage I translates the given human pose to the human parsing according to the text describing the clothes shapes. The text for clothes shapes is first transformed to one-hot shape attributes and embedded to a vector $f_{shape}$.
   The shape vector $f_{shape}$ is then fed into the pose-to-parsing module to spatially modulate the pose features. Stage II generates the human image from the synthesized human parsing by sampling multi-level indices from our learned hierarchical texture-aware codebooks.
   To sample coarse-level indices, 
   we employ a sampler with mixture-of-experts, where features are routed to different expert heads to predict the indices based on the required textures. At the fine level, we propose a feed-forward network to efficiently predict fine-level indices to
   refine the generated human image.}
   \label{pipeline_illustration}
\end{figure*}

To summarize, our main contributions are as follows:
\textbf{1)} We propose the Text2Human framework for the task of text-driven controllable human generation. Our proposed framework is able to generate photo-realistic human images from natural language descriptions.
\textbf{2)} We build a hierarchical VQVAE with the texture-aware codebook design. We propose a transformer-based sampler with the concept of mixture-of-experts. The features are routed to different expert heads according to the required attributes. The hierarchical design and mixture-of-experts sampler enable the synthesis and control of complicated textures.
\textbf{3)} We propose a feed-forward index prediction network to predict codebook indices of fine-level codebook based on the features sampled at the coarse level, which overcomes the limitation of the time-consuming sampling process in classical hierarchical VQVAE methods.
\textbf{4)} We contribute a large-scale and high-quality human image dataset with rich clothes shape and texture annotations as well as human parsing masks to facilitate the task of controllable human synthesis.

\section{Related Work}

\paragraph{Generative Models.} 
Generative Adversarial Network (GAN) has demonstrated its powerful capabilities in generating high-fidelity images. 
Since \cite{goodfellow2014generative} proposed the first generative model in 2014, different variants of GAN \cite{brock2018large, karras2020analyzing, karras2021alias, karras2019style,chai2022any} have been proposed.
In addition to unconditional generation, conditional GANs \cite{mirza2014conditional} were proposed to generate images based on conditions like segmentation mask \cite{isola2017image, wang2018high, park2019semantic} and natural language \cite{xu2018attngan, surya2020restgan}. 
Our proposed Text2Human is a conditional image generation framework by taking human poses and texts as inputs.
In parallel to GAN, VAE \cite{kingma2013auto} is another paradigm for image generation. It embeds input images into a latent distribution and synthesizes images by sampling vectors from the prior distribution.
Several VAE-based works \cite{larsen2016autoencoding, esser2018variational, oord2017neural, esser2021taming} have been proposed to improve the visual quality of the generated images.
Our proposed method shares some similarities with existing VAE-based methods but differs in the texture-aware codebook, sampler with mixture-of-experts, and feed-forward index prediction network for the hierarchical sampling.

\paragraph{Human Image Manipulation and Synthesis.}
The goal of pose transfer \cite{ma2017pose, ma2018disentangled, liu2019neural, liu2020neural, balakrishnan2018synthesizing,tao2022structure} is to transfer the appearance of the same person from one pose to another. \cite{albahar2021pose} proposed a pose-conditioned StyleGAN framework. The details of the source image are warped to the target pose and then
are used to spatially modulate the features for synthesis. 
\cite{zhou2019text} proposed a method for the text-guided pose transfer task.
\cite{men2020controllable} proposed ADGAN for controllable person image synthesis. The person image is synthesized by providing a pose and several example images.
All of these tasks require a source person image to synthesize the target person.
Recently, TryOnGAN \cite{lewis2021tryongan} and HumanGAN \cite{sarkar2021humangan} are proposed to support the human image generation conditioned on human pose only.
TryOnGAN trained a pose conditioned StyleGAN2 network and can generate human images under the given pose condition. HumanGAN proposed a VAE-based human image generation framework. Human images are generated by sampling from the learned distribution.
However, these methods do not offer fine-grained controls on human generation. Our proposed framework allows for controllable human generation by giving texts describing the desired attributes.

\section{Text2Human}

\begin{figure}
   \begin{center}
      \includegraphics[width=0.96\linewidth]{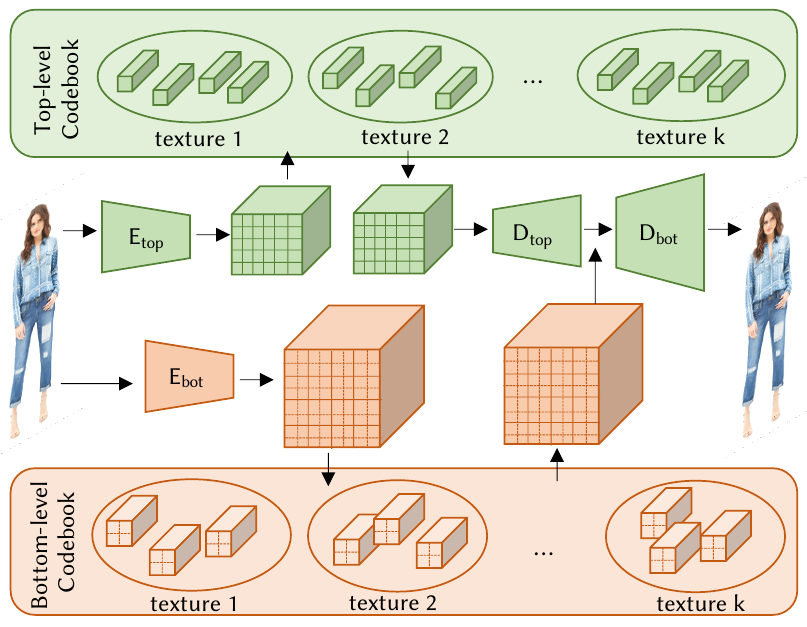}
   \end{center}
   \caption{\textbf{Illustration of Hierarchical VQVAE and Texture-Aware Codebooks.} The images are reconstructed using two levels of features, \ie, top level for coarse-scale features and bottom level for fine-scale features. Texture-Aware Codebooks are built for different types of clothing textures.}
   \label{hier_vqvae}
\end{figure}

Our aim is to generate human images conditioned on texts describing the attributes of clothes (clothes shapes and clothes textures).
Given a human pose $P \in \mathbb{R}^{H \times W}$, texts for clothes shapes $T_{shape}$, and texts for clothes textures $T_{texture}$, the output should be the corresponding human image $I \in \mathbb{R}^{H \times W \times 3}$.
The whole pipeline of Text2Human is shown in Fig.~\ref{pipeline_illustration}. We decompose the human generation into two stages. Stage I synthesizes a human parsing mask with the given pose and texts for clothes shapes.
We transform the text information to attribute embeddings and concatenate them with human pose features to predict the desired human parsing mask.
With the human parsing mask obtained from Stage I as the input, the final image is synthesized according to the required clothing textures in Stage II. 
We set up a hierarchical texture-aware codebook to characterize various types of texture as illustrated in Fig.~\ref{hier_vqvae}, where the final image is synthesized using both coarse-level (top-level) and fine-level (bottom-level) codebooks.
To sample the codebook indices at the coarse level, a sampler with mixture-of-experts is proposed, where features are routed to different expert heads to predict the desired indices. To speed up the sampling at the fine level, we propose a feed-forward codebook index prediction network, which further refines the quality of generated images.

\subsection{Stage I: Pose to Parsing}
Given a human pose $P$ and texts about clothes shapes,
we hope to synthesize the human parsing map $S \in \mathbb{R}^{H \times W}$. 

First, texts are transformed to a set of clothes shape attributes $\{a_1, ..., a_i, ..., a_k\}$, where $a_i \in \{0, 1, ..., C_i\}$ and $C_i$ is the class number of attribute $a_i$. The attributes are then fed into the Attribute Embedding Module to obtain a shape attribute embedding $f_{shape} \in \mathbb{R}^C$: 
\begin{equation}
    f_{shape} = Fusion([E_1(a_1), E_2(a_2), ..., E_i(a_i), ..., E_k(a_k)]),
\end{equation}
where $E_i(\cdot)$ is the attribute embedder for $a_i$ and $Fusion(\cdot)$ fuses attribute embeddings from $k$ attribute embedders. $[\cdot]$ denotes the concatenation operation. 

Together with $P$, the $f_{shape}$ is then fed into the Pose-to-Parsing Module, which is composed of an encoder $Enc$ and a decoder $Dec$.
The operation at layer $i$ of $Enc$ is defined as follows:
\begin{equation}
    f_{p_i} = Enc_i([f_{p_{i-1}}, \mathcal{B}(f_{shape})]),
\end{equation}
where $\mathcal{B}(\cdot)$ is the spatial broadcast operation so that $f_{shape}$ is broadcasted to have the same spatial size with $f_{p_{i-1}}$, and $f_{p_0} = P$.

The operation of $Dec$ at layer $i$ can be expressed as $f_{p_i}^{\prime} = Dec_i([f_{p_i},$ $f_{p_{i-1}}^{\prime}])$.
The final decoded feature $f_{p}^{\prime}$ is fed into fully convolutional layers to make the final parsing prediction. We use the cross-entropy loss to train the whole Pose-to-Parsing Module.

\subsection{Stage II: Parsing to Human}

\subsubsection{Preliminaries}
\paragraph{VQVAE} The goal of Vector-Quantized Variational AutoEncoder (VQVAE) \cite{oord2017neural} is to learn a discrete codebook that stores discrete neural representations by learning to reconstruct images.
VQVAE consists of an encoder $E$, a decoder $G$ and a learnable codebook $\mathcal{Z} = \{ z_k | z_k \in \mathbb{R}^{c_z}\}_{k=1}^{K}$. We first extract the continuous neural representation $\hat{z}$ by feeding the image $I$ into the encoder, \ie, $\hat{z} = E(I) \in \mathbb{R}^{h \times w \times c_z}$. Then the quantizer $Quant$ is adopted to discretize the continuous $\hat{z}$, and the operation is defined as follows:
\begin{equation}
    z_q = Quant(\hat{z}) := \underset{z_k \in \mathcal{Z}}{\mathrm{argmin}} \left\| \hat{z}_{ij} - z_k \right\| \in \mathbb{R}^{h \times w \times c_z}. 
\label{eq:quant}
\end{equation}
Then the image is reconstructed using the quantized representation $\hat{I} = G(z_q)$. The encoder, decoder and codebook are end-to-end trained through the following loss function:
\begin{equation}
    \mathcal{L} = \left\| I - \hat{I} \right\| + \left\| sg(\hat{z}) - z_q \right\|_2^2 + \left\| sg(z_q) - \hat{z} \right\|_2^2,
\label{eq:loss_vqvae}
\end{equation}
where $sg(\cdot)$ denotes the stop-gradient operation.

\paragraph{Diffusion-based Transformer.}
To sample images from learned codebooks, autoregressive models \cite{salimans2017pixelcnn++, chen2018pixelsnail} are employed to predict the orderings of codebook indices.
Autoregressive models predict indices in a fixed unidirectional manner and the prediction of the incoming index only relies on already sampled top-left parts.
In VQVAE, PixelCNN \cite{oord2016conditional} is adopted as the autoregressive model. In recently proposed VQGAN \cite{esser2021taming}, transformer \cite{vaswani2017attention} is adopted for its capability to capture long-term dependencies among codebook indices (In transformer, codebook indices are referred to as `tokens').
Recently, some works \cite{bond2021unleashing, gu2021vector, esser2021imagebart,chang2022maskgit} proposed to use the diffusion model to replace the autoregressive model motivated by two advantages: 1) Indices are predicted based on global and bidirectional context, resulting in more coherent sampled images; 2) Indices are predicted in parallel, leading to much faster sampling speed.
Specifically, 
in diffusion-based transformer, starting from fully-masked indices $k_0$, the final prediction of indices $k_T$ are sampled $T$ steps by transformers. The indices $k_t$ at the step $t$ are sampled following the distributions:
\begin{equation}
    k_t \sim q_{\theta}(k_t | k_{t-1}),
\end{equation}
where $\theta$ is the parameters of transformers.
At each time step, the indices are randomly replaced with newly sampled ones. 

\subsubsection{Hierarchical VQVAE with Texture-Aware Codebook}
%
Considering the complicated nature of clothes textures, representing textures in single-scale features is not enough.
For example, as shown in Fig.~\ref{fig:ablation}(a), the reconstruction of a plaid shirt with multi-scale features contains more details.
Inspired by this, we propose the hierarchical VQVAE with multi-scale codebooks. Specifically, given an input image $I \in \mathbb{R}^{H \times W \times 3}$, we first train an encoder $E_{top}$ to downsample $I$ to obtain its coarse-level feature $\hat{feat}_{top}$:
\begin{equation}
    \hat{feat}_{top} = E_{top}(I) \in \mathbb{R}^{H/16 \times W/16 \times c_z}.
\end{equation}

We build a top-level codebook $\mathcal{Z}_{top}$ for $\hat{feat}_{top}$ with codes $\in\mathbb{R}^{1 \times 1 \times c_z}$.
The quantization of $\hat{feat}_{top}$ is the same as Eq.~(\ref{eq:quant}).

Then the image is reconstructed using the quantized feature $feat_{top}$ through the decoder $D$: $\hat{I} = D(feat_{top})$. Here we view $D$ as two consecutive parts $D = D_{bot} \circ D_{top}$. The spatial sizes of the inputs to $D_{bot}$ and $D_{top}$ are $H/8 \times W/8$ and $H/16 \times W/16$, respectively.

Once the top-level codebook $\mathcal{Z}_{top}$ is trained, we move to build the bottom-level codebook $\mathcal{Z}_{bot}$.
The image features represented by the codes of $\mathcal{Z}_{top}$ already recover the coarse information. Therefore, $\mathcal{Z}_{bot}$ just needs to learn residual information to $\mathcal{Z}_{top}$. 
We introduce a residual encoder $E_{bot}$ to extract 
fine-level feature $\hat{feat_{bot}}$, which is quantized into $feat_{bot}$ with $\mathcal{Z}_{bot}$.
The image is then constructed as follows:
\begin{equation}
    \hat{I} = D_{bot}(D_{top}(feat_{top}) + feat_{bot}).
\end{equation}
During the training of bottom-level codebook and $E_{bot}$, $E_{top}$ and $D_{top}$ are fixed. The network is optimized by Eq.~(\ref{eq:loss_vqvae}) combined with the perceptual loss and discriminator loss.

To make the codes in $\mathcal{Z}_{bot}$ contain richer texture information as well as keep the well-learned structure information in $\mathcal{Z}_{top}$, the code shape is set to $2 \times 2 \times c_z$ rather than the conventional $1 \times 1 \times c_z$.
It is implemented by dividing $feat_{bot}$ into non-overlapping patches with spatial size of $2 \times 2$. Once the features are divided into patches, the quantization process is the same as Eq.~(\ref{eq:quant}).

Our hierarchical VQVAE shares some similarities with VQVAE2 \cite{razavi2019generating} in the hierarchical design, but differs in the following aspects: 1) Codes in our fine-level codebook have a spatial size of $2\times2$, while the codes in codebooks of VQVAE2 has no spatial size; 2) Our hierarchical design is motivated by representing textures at multiple scales while VQVAE2 is motivated to learn more powerful priors over the latent codes. 3) VQVAE2 trains the whole network end-to-end, which leads to poor representation ability of coarse-level features. Our stage-wise training strategy ensures meaningful representations at all levels.

Apart from multi-level codebooks, we further design a texture-aware codebook. The motivation behind the texture-awareness of the codebook lies in that the textures with different appearances at the original scale may appear to be similar at downsampled scales, leading to an ambiguity problem if we build a single coarse-level codebook for all textures. 
Therefore, we build different codebooks for different texture attributes separately.
We will divide features extracted by the encoders according to their texture attributes at the image level and feed them into different codebooks to get the quantized features.

\subsubsection{Sampler with Mixture-of-Experts}

To incorporate texture-aware codebooks, we adapt the diffusion-based transformer into a texture-aware one as well.
A straightforward idea is to train multiple samplers for different textures. However, this naive idea has two shortcomings: 
1) Contextual information in the whole image is vital for the sampling of codebook indices, while training sampler for one single texture makes such information blind to the network. 
2) Training multiple samplers are not ideal if we adopt the transformer as the sampler, since multiple transformers are too heavy for modern GPU devices.

Therefore, we introduce the idea of mixture-of-experts \cite{shazeer2017outrageously} into the diffusion-based transformer.
The inputs to the mixture-of-experts sampler consist of three parts: 1) codebook index $T_{code}$, 2) tokenized human segmentation masks $T_{seg}$, and 3) tokenized texture masks $T_{tex}$. The texture mask is obtained by filling the texture attribute labels of clothes in the corresponding regions of the segmentation mask. 
The multi-head attention $MHA(\cdot)$ of the transformer is computed among all of the tokens:
\begin{equation}
    f = MHA(Emb_{code}(T_{code}) + Emb_{seg}(T_{seg}) + Emb_{tex}(T_{tex})),
\end{equation}
where $Emb_{code}$, $Emb_{seg}$ and $Emb_{tex}$ are learnable embeddings.

The feature $f$ extracted by the multi-head attention is routed to different experts heads. The router routes the specific textures based on the texture attribute information provided by $T_{texture}$. Each expert head is in charge of the prediction of tokens for a single texture. The prediction of tokens is formulated as a classification task, where the class number is the size of the codebook.
The final codebook indices are composed of outputs from all expert heads.

During training, the codebook index $T_{code}$ is the coarse-level codebook index obtained by the hierarchical VQVAE. When it comes to sampling, $T_{code}$ is initialized with masked tokens and it is iteratively filled with newly sampled ones until fully filled.

\subsubsection{Feed-forward Codebook Index Prediction}

To sample an image from the hierarchical VQVAE, multiple feature maps composed of the hierarchical codebooks need to be fed into the decoders. The traditional paradigm \cite{razavi2019generating} is to sample multiple features at different scales. However, token-wisely sampling at larger feature scales is time-consuming. Besides, when sampling at a large feature scale, long-term dependencies are hard to capture, and thus the generated images are of poor quality. 

Motivated by these, we propose a feed-forward codebook index prediction network by harnessing the implicit relationship between codebooks at different levels learned by our proposed hierarchical VQVAE.
Specifically, features, which are token-wisely sampled at the coarse level, are fed into the codebook index prediction network to predict the fine-level codebook indices.
The codebook index prediction network $N$ is defined as:
\begin{equation}
    index_{bot} = N(feat_{top}).
\end{equation}
The encoder-decoder network is adopted for the index prediction network. 
It should be noted that the codebook index prediction network is texture-aware as well. Shared features are extracted by the encoder and decoder, but fed into different classifier heads according to the attributes.

The use of the codebook index prediction network and the hierarchical codebooks improves the quality of generated images compared to images generated with only one level codebook. Thanks to the feed-forward index prediction network, the sampling process at larger scales under the hierarchical VQVAE design can be achieved within only one single forward pass. It speeds up the sampling process compared to the token-wisely autoregressive sampling used in \cite{razavi2019generating}.

\subsection{Text-driven Synthesis}
Our framework is a text-driven one. To transform the texts requested by users into attributes,
we have some predefined text descriptions for each attribute.
We use the pretrained Sentence-BERT model \cite{reimers-2019-sentence-bert} to extract the word embeddings of our predefined texts and the text requested by users and then calculate their cosine similarities. According to the cosine similarities of word embeddings, we then classify the texts into their corresponding attributes.

\subsection{Interactive User Interface}

\begin{figure}
  \begin{center}
      \includegraphics[width=0.96\linewidth]{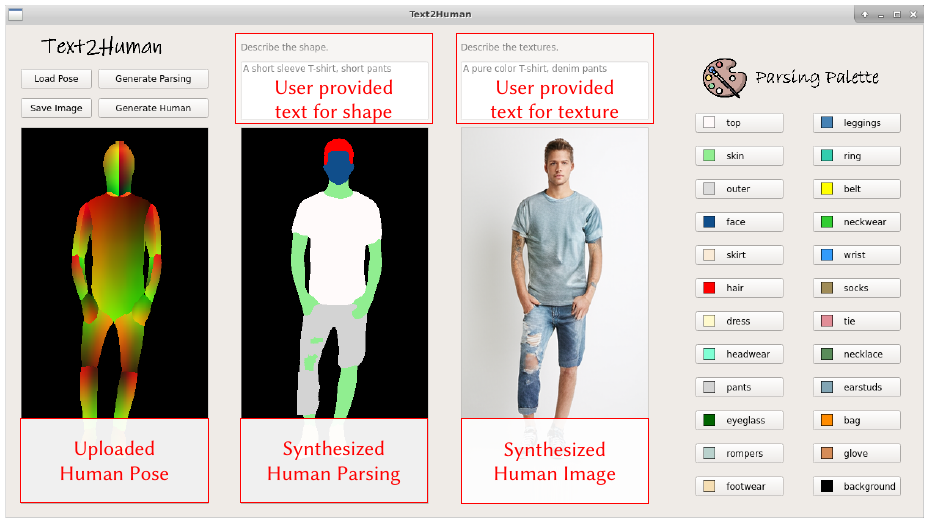}
  \end{center}
  \caption{\textbf{User Interface with Parsing Palette.} To generate the human image, users are required to upload a human pose and texts describing the clothing shapes and textures. Users can modify the generated human parsing by using the parsing palette. For example, they can edit the right pant leg from a short one to a long one. Some holes can be added to the right pant leg to make the results more customized.}
  \label{fig:ui}
\end{figure}

We present an interactive user interface for our Text2Human as shown in Fig.~\ref{fig:teaser}(a).
Users can upload a human pose map and then type a text describing the clothing shapes. A human parsing map will be generated accordingly. Then users provide another text describing the clothing textures, and Text2Human generates the corresponding final human image.
On the right side of the interface, we provide a parsing palette, which enables users to edit the human parsing. For example, as shown in Fig.~\ref{fig:ui}, users can draw some holes on jeans and make the right pant leg longer using the palette to make the generated images more customized. 

\section{DeepFashion-MultiModal Dataset}

Currently, most human generation methods are developed on the low-resolution version of the DeepFashion dataset and the datasets lack fine-grained annotations. 
Therefore, a publicly available and well-annotated high-quality human image dataset is important for the research on the human generation task.
Motivated by this, we set up a large-scale high-quality human dataset with rich attribute annotations named DeepFashion-MultiModal Dataset. In a nutshell, our dataset has the following properties: 1) It contains 11,484 high-quality images at $1024 \times 512$ resolution. 2) For each image, we manually annotate the human parsing labels with 24 classes. 3) Each image is annotated with attributes for both clothes shapes and textures. 4) We provide densepose for each human image.

\paragraph{Data Source and Processing.}
DeepFashion dataset is a large-scale clothes database that contains over 800,000 fashion images, ranging from in-shop images to unconstrained photos uploaded by customers on e-commerce websites with varying quality.
Since images from the in-shop clothes retrieval benchmark are mostly of high quality with pure color background, we filter full-body images from this benchmark. There are 11,484 full-body images in total.
Similar to the data alignment method used in FFHQ~\cite{karras2019style}, we align the full-body images based on their poses.

\paragraph{Annotations.}
\textbf{1)} Human Pose Representations:
We extract densepose for each image using the off-the-shelf method~\cite{guler2018densepose}. 
\textbf{2)} Human Parsing Annotations:
Human parsing serves as an effective intermedium in pose-to-photo synthesis. For each image, we provide human parsing annotations including 24 semantic labels of body components (face, hair, skin), 
clothes (top, outer, skirt, dress, pants, rompers) and accessories (headwear, eyeglasses, neckwear, \textit{etc}.).
The human parsing is manually annotated from scratch by annotators using Photoshop.
\textbf{3)} Clothes Shape Annotations:
We manually label the clothes shape attributes for each image. The annotations include the length of upper clothes and lower clothes, the presence of fashion accessories (\eg, hat, glasses, neckwear), and the shapes of the upper clothes' necklines. The length of upper clothes falls into four classes: sleeveless, short-sleeve, medium-sleeve, and long-sleeve. The categories for lower clothes are three-point shorts, shorts, cropped pants, and trousers. The shapes of necklines are roughly divided into V-shape, square-shape, crew neck, turtleneck, and lapel. The presence of fashion accessories has two states, \textit{i.e.}, presence or absence. 
When we annotate clothes shapes for jumpsuits (\eg, dress and rompers), the upper part and the lower part of garments are treated separately.
\textbf{4)} Clothes Texture Annotations:
We manually label the clothes textures by two orthogonal dimensions: clothes colors and clothes fabrics.
Clothes colors consist of floral, patterned, stripes, solid color, lattice, color blocks, and hybrid colors.
Clothes fabrics are divided into denim, cotton, leather, furry, knitted, tulle, and other materials.

\section{Experiments}

\subsection{Implementation Details}
We split the dataset into a training set and a testing set. The training set contains $10,335$ images and the testing set contains $1,149$ images.
We downsample the images to $512 \times 256$ resolution.
The texture attribute labels are the combinations of clothes colors and fabrics annotations.
The modules in the whole pipeline are trained stage by stage.
All of our models are trained on one NVIDIA Tesla V100 GPU. We adopt the Adam optimizer. The learning rate is set as $1 \times 10^{-4}$.
For the training of Stage I (\ie, Pose to Parsing), we use the (human pose, clothes shape labels) pairs as inputs and the labeled human parsing masks as ground truths. We use the instance channel of densepose (three-channel IUV maps in original) as the human pose $P$. Each shape attribute $a_i$ is represented as one-hot embeddings. We train the Stage I module for $50$ epochs. The batch size is set as $8$.
For the training of hierarchical VQVAE in Stage II, we first train the top-level codebook, $E_{top}$, and decoder for 110 epochs, and then train the bottom-level codebook, $E_{bot}$, and $D_{bot}$ for 60 epochs with top-level related parameters fixed. The batch size is set as $4$. 
The sampler with mixture-of-experts in Stage II requires $T_{seg}$ and $T_{tex}$. $T_{seg}$ is obtained by a human parsing tokenizer, which is trained by reconstructing the human parsing maps for $20$ epochs with batch size $4$. $T_{tex}$ is obtained by directly downsampling the texture instance maps to the same size of codebook indices maps using nearest interpolation. The cross-entropy loss is employed for training. The sampler is trained for $90$ epochs with the batch size of $4$.
For the feed-forward index prediction network, we use the top-level features and bottom-level codebook indices as the input and ground-truth pairs. The feed-forward index prediction network is optimized using the cross-entropy loss.
The index prediction network is trained for $45$ epochs and the batch size is set as $4$.

\begin{table}[]
\begin{center}
\caption{\textbf{Quantitative Comparisons} on human images generated given human parsing maps and clothes textures. Our method achieves the lowest FID score and highest attribute prediction accuracy for complicated textures.}
\begin{tabular}{l|ccccc}
\Xhline{1pt}
\textbf{Methods}    & \textbf{FID}  & \textbf{Floral} & \textbf{Stripe} & \textbf{Denim} \\ \Xhline{1pt}
Pix2PixHD      & 39.80 & 52.94\% & 69.44\% & \textbf{99.03\%} \\ \hline
SPADE        & 30.13  & 61.76\% & 55.56\% & 98.55\% \\ \hline
MISC  & 27.97  & 50.00\% & 69.44\% & 97.58\% \\ \hline 
HumanGAN-parsing & 27.71  & 8.82\% & 5.56\% & 87.17\% \\ \hline
Text2Human-parsing & \textbf{22.95} & \textbf{70.59\%} & \textbf{88.89\%} & 95.88\% \\
\Xhline{1pt}
\end{tabular}
\label{quant:parsing}
\end{center}
\end{table}

\begin{table}[]
\begin{center}
\caption{\textbf{Quantitative Comparisons} on human images generated given human poses. Our method achieves the lowest FID score and the largest ratio of floral, stripe, and lattice textures on clothes.}
\begin{tabular}{l|ccccc}
\Xhline{1pt}
\textbf{Methods}    & \textbf{FID}  & \textbf{Floral}  & \textbf{Stripe } & \textbf{Lattice} \\ \Xhline{1pt}
HumanGAN-pose & 32.20 & 2.00\% & 0.61\% & 0.35\% \\ \hline
TryOnGAN & 29.00 & 3.05\% & 2.52\% & 0.44\% \\ \hline
Text2Human-pose & \textbf{24.54} & \textbf{3.57\%} & \textbf{4.61\%} & \textbf{1.39\%} \\
\Xhline{1pt}
\end{tabular}
\label{quant:pose}
\end{center}
\end{table}

\subsection{Comparison Methods}
\paragraph{Pix2PixHD} \cite{wang2018high} is a conditional GAN for semantic map guided image synthesis. Here, we use the human parsing map and the texture map obtained by filling texture attribute labels in the human parsing map as inputs.

\paragraph{SPADE} \cite{park2019semantic} is a conditional GAN for semantic map guided synthesis. It is adapted in a similar way to Pix2PixHD.

\paragraph{MISC} \cite{weng2020misc} synthesizes human images based on a human parsing map and some attributes about the clothes. 

\paragraph{HumanGAN} \cite{sarkar2021humangan} is a pose-conditioned VAE-based human generation method, which generates diverse human appearances by sampling from a fixed distribution (\eg, Gaussian distribution). 

\paragraph{TryOnGAN} \cite{lewis2021tryongan} is a pose-conditioned StyleGAN method. The constant noise is replaced with the pose features. 
We train the model with the same human pose representation as our method for fair comparisons.

\paragraph{Taming Transformer} \cite{esser2021taming} is a VQVAE-based method and also shows an application to conditional human image generation. For a fair comparison, we use human parsing as the input condition.

\subsection{Evaluation Metrics}
\paragraph{FID} 
For image generation tasks, Fréchet Inception Distance (FID) is a metric evaluating the similarities between generated images and training images. A lower FID indicates a higher quality. 

\paragraph{Attribute Prediction Accuracy.} 
We use a pretrained predictor to predict the texture attributes of generated images. The prediction accuracy is reported to measure the realism of the generated texture. We also use the pretrained predictor to calculate the ratios of complicated textures (floral, stripe, lattice) to evaluate the diversity.

\paragraph{User Study.} 
A user study is performed to evaluate the quality of the generated images. Users are presented with 20 groups of results. Each group has five images generated by baselines and our method. A total of 16 users are asked to 1) rank images according to photorealism (rank 5 is the best) and 2) score texture consistency with the given three attribute labels for upper clothes, lower clothes and outer clothes. The full score is 3. If the outer clothing is not required, the score for the outer clothing is 1.

\begin{figure}
   \begin{center}
      \includegraphics[width=1.0\linewidth]{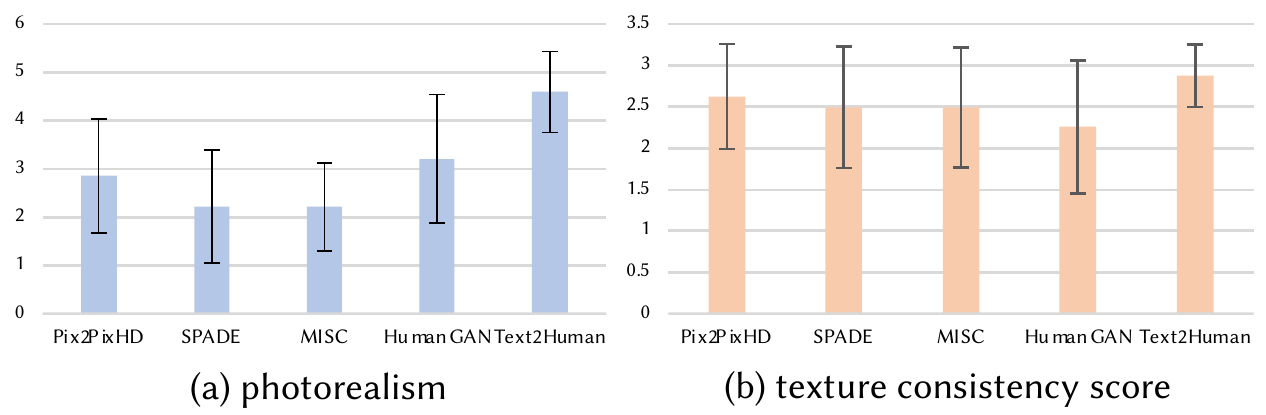}
   \end{center}
   \caption{\textbf{User Study Results.} (a) Average rank for the photorealism of generated images. A higher rank indicates a better visual quality. (b) Scores of the consistency of the textures of generated images and provided labels.}
   \label{fig:user_study}
\end{figure}

\subsection{Quantitative Comparisons}
We report the quantitative results under two different settings: human image generation 1) from a human parsing, and 2) from a given human pose.
Table~\ref{quant:parsing} shows the comparisons with state-of-the-art conditional image generation methods. A well-annotated human parsing map and labels for clothes texture annotations are provided to synthesize the human images. As shown in Table~\ref{quant:parsing}, our method achieves the lowest FID, which demonstrates the fidelity and diversity of our generated human images. 
In addition, the best texture attribute prediction accuracy shows that our proposed Text2Human framework can accurately generate human images conditioned on provided textures.
In Table~\ref{quant:pose}, we show the quantitative comparisons on pose-guided human image synthesis. Since it is non-trivial to add clothes shape and texture controls for HumanGAN and TryOnGAN, under this setting, 
we report the ratio of complicated textures among all generated images. The highest ratio demonstrates that our methods can synthesize diverse textures for clothes.
The user study results are shown in Fig.~\ref{fig:user_study}. Our proposed Text2Human gets the highest rank in terms of the photorealism of the generated images. As for the clothes textures, the images synthesized by our framework are more consistent with the required texture attributes.
The user study results are consistent with other quantitative results.

\subsection{Qualitative Comparisons}

\begin{figure}
  \begin{center}
      \includegraphics[width=1.0\linewidth]{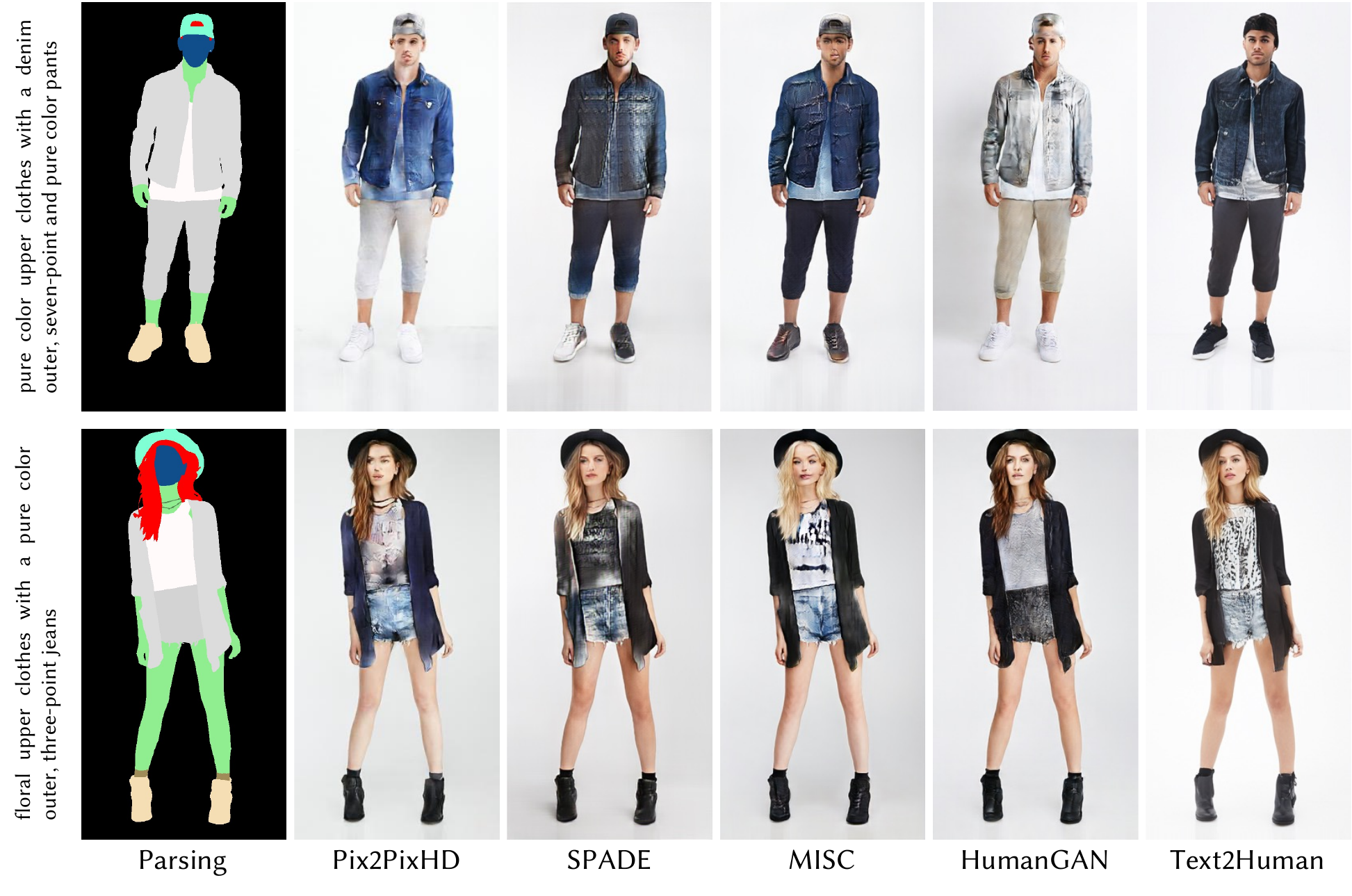}
  \end{center}
  \caption{\textbf{Qualitative Comparison} on image generation given human parsing maps and clothes textures. Our proposed method can generate complicated textures with finer details and high-fidelity faces.} 
  \label{fig:compare1}
\end{figure}

\begin{figure}
  \begin{center}
      \includegraphics[width=1.0\linewidth]{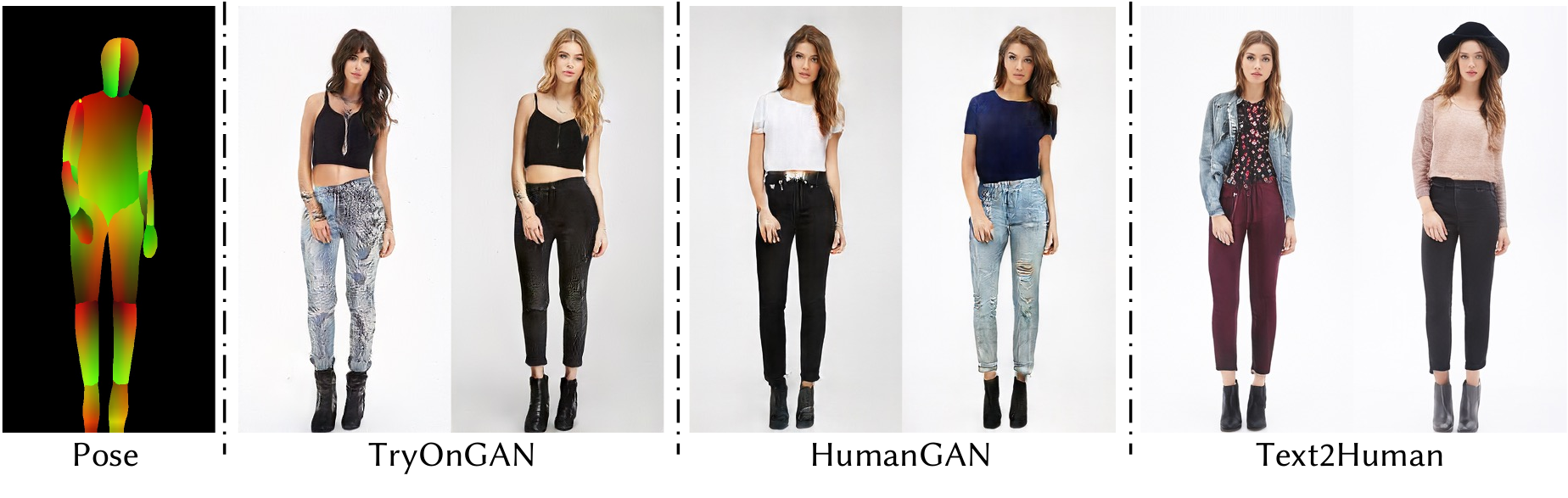}
  \end{center}
  \caption{\textbf{Qualitative Comparison} on Pose-guided Image Generation. The compared baselines do not offer any controls on clothes shapes and textures, while our method can explicitly control these attributes.}
  \label{fig:compare2}
\end{figure}

\begin{figure}
  \begin{center}
      \includegraphics[width=1.0\linewidth]{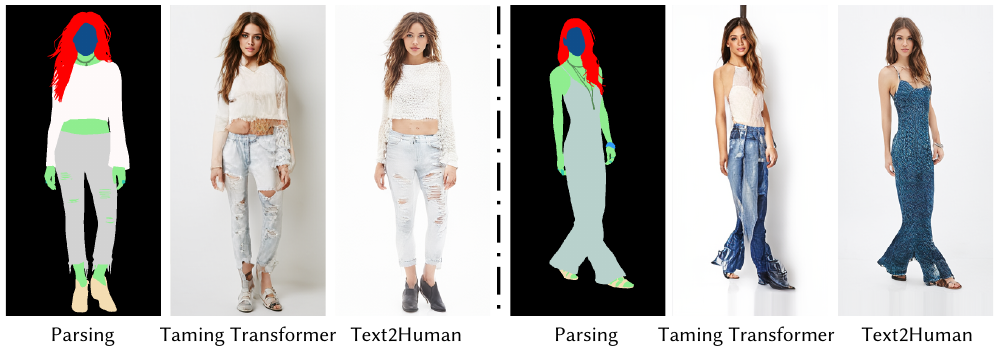}
  \end{center}
  \caption{\textbf{Qualitative Comparison} with Taming Transformer. Given the human parsing as input, our method can generate human images with better fidelity.}
  \label{fig:compare3}
\end{figure}

Figure~\ref{fig:compare1} shows visual comparisons on synthesized human images given human parsing maps and clothes textures. Our proposed method can generate complicated textures with finer details and high-fidelity faces.
Figure~\ref{fig:compare2} shows visual comparisons with state-of-the-art pose-guided TryOnGAN~\cite{lewis2021tryongan} and HumanGAN~\cite{sarkar2021humangan}. The compared baselines do not offer any controls on clothes shapes and textures, while our method can explicitly control these attributes.
We also compare our proposed Text2Human with another VQVAE-based method, Taming Transformer \cite{esser2021taming}. As shown in Fig.~\ref{fig:compare3}, given the same human parsing map, our method can generate more plausible human images.

\subsection{Ablation Study}
\begin{figure*}
  \begin{center}
      \includegraphics[width=0.9\linewidth]{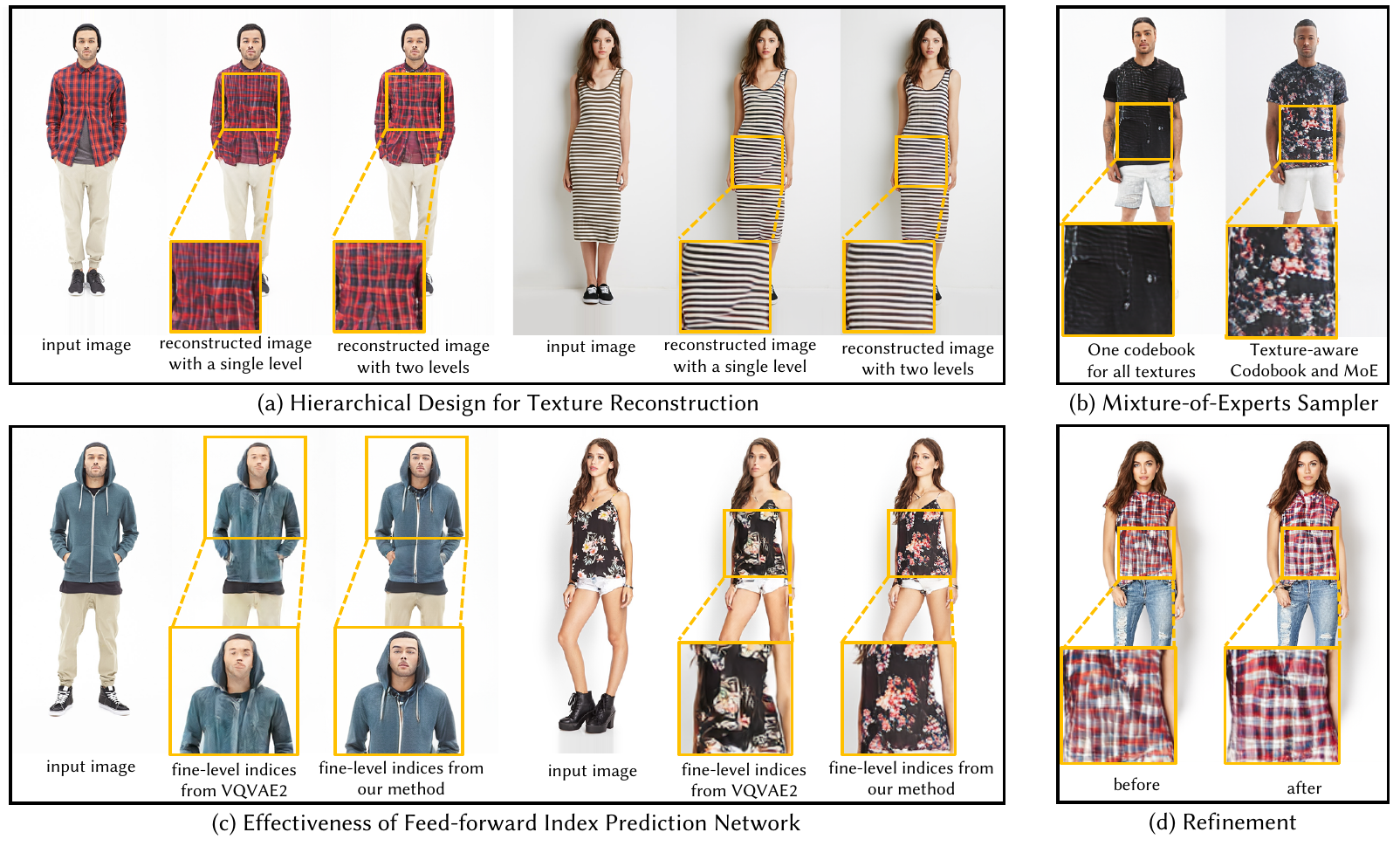}
  \end{center}
  \caption{\textbf{Ablation Study.} (a) The hierarchical design recovers more high-frequency details. (b) Mixture-of-Experts sampler enables the framework to synthesize images conditioned on requested texture inputs. (c) Compared to VQVAE2, our feed-forward network generates clearer images. (d) With the feed-forward network, sampled lattice patterns are refined.}
  \label{fig:ablation}
\end{figure*}

\paragraph{Hierarchical Design for Texture Reconstruction.} 
Fig.~\ref{fig:ablation}(a) shows the improvement brought by the proposed hierarchical VQVAE on the recovery of plaid and stripe patterns. With the hierarchical design, the reconstructed images contain more high-frequency details, verifying better texture representations are learned.
The hierarchical design reduces the reconstruction loss (\ie, $l_{1}$ loss + perceptual loss) from $0.1415$ to $0.1192$ on the whole testing set.

\begin{table}[]
\begin{center}
\caption{\textbf{Quantitative Results} on the effectiveness of texture-aware codebook and mixture-of-experts. We report the attribute prediction accuracy for models with MoE and without MoE.}
\begin{tabular}{l|ccc}
\Xhline{1pt}
\textbf{Methods} & \textbf{Floral}  & \textbf{Stripe} & \textbf{Denim} \\ \Xhline{1pt}
Without MoE & 20.59\% & 22.22\% & 92.01\% \\ \hline
With MoE &  \textbf{70.59\%} & \textbf{88.89\%} & \textbf{95.88\%} \\
\Xhline{1pt}
\end{tabular}
\label{quant:moe}
\end{center}
\end{table}

\begin{table}[]
\begin{center}
\caption{\textbf{LPIPS distance and ArcFace cosine distance} between the reconstructed images and original images.}
\begin{tabular}{l|cc}
\Xhline{1pt}
\textbf{Methods} & \textbf{LPIPS $\downarrow$}  & \textbf{ArcFace $\uparrow$} \\ \Xhline{1pt}
VQVAE2 & 0.0791 & 0.3415 \\ \hline
Ours &  \textbf{0.0609} & \textbf{0.4869} \\
\Xhline{1pt}
\end{tabular}
\label{quant:prediction}
\end{center}
\end{table}

\paragraph{Texture-Aware Codebook and Mixture-of-Experts Sampler.} 
To evaluate the effectiveness of our texture-aware and mixture-of-experts design, we train a diffusion-based sampler with only one codebook for all textures. 
As shown in Fig.~\ref{fig:ablation}(b), the sampler without mixture-of-experts and texture-aware codebook cannot generate requested floral textures, demonstrating our design makes the sampler better conditioned on the textual inputs.
We report attribute prediction accuracies on complicated textures (\ie, floral, stripe, and denim). The results are shown in Table~\ref{quant:moe}. Without mixture-of-experts, the attribute prediction accuracy drops by $50.00\%$, $66.67\%$, and $3.87\%$ on floral, stripe, and denim textures, respectively. 
There are more denim textures (3449 images) than floral (325 images) and stripe (361 images) textures in the training set. It is easier for models to capture the patterns of denim textures even without the mixture-of-expert design. As a result, we can observe a smaller performance gap for denim textures, compared to those for floral and stripe textures. It indicates that the mixture-of-experts design is more effective in generating uncommon textures with fewer training samples.

\paragraph{Feed-Forward Index Prediction Network.} 
To overcome the limitations of the hierarchical sampling paradigm of VQVAE2, we propose a feed-forward index prediction network to speed up the sampling speed as well as refine the textures. 
In terms of running time, our feed-forward network predicts fine-level codebook indices within \textbf{0.6s} while VQVAE2 takes \textbf{25mins}.
In terms of quality, we conduct a comparative experiment with VQVAE2. 
For a fair comparison, we use the ``ground-truth'' coarse-level codebook indices obtained when reconstructing a given human image as input to predict fine-level indices by the autoregressive model of VQVAE2 or our feed-forward network. 
As shown in Fig.~\ref{fig:ablation}(c), our method reconstructs more clear and high-fidelity clothes textures than VQVAE2.
We report LPIPS distance \cite{zhang2018perceptual} and ArcFace distance \cite{deng2019arcface} between the reconstructed images and the original images in Table~\ref{quant:prediction}. It further verifies the effectiveness of our proposed feed-forward index prediction network in terms of reconstruction performance.
Fig.~\ref{fig:ablation}(d) further provides a visualization of the refinement of our feed-forward network. Our network effectively refines the synthesized lattice patterns sampled from the coarse-level codebook.

\subsection{Limitations}
In this section, we discuss three common limitations of our proposed Text2Human.
\begin{figure}
  \begin{center}
      \includegraphics[width=1.0\linewidth]{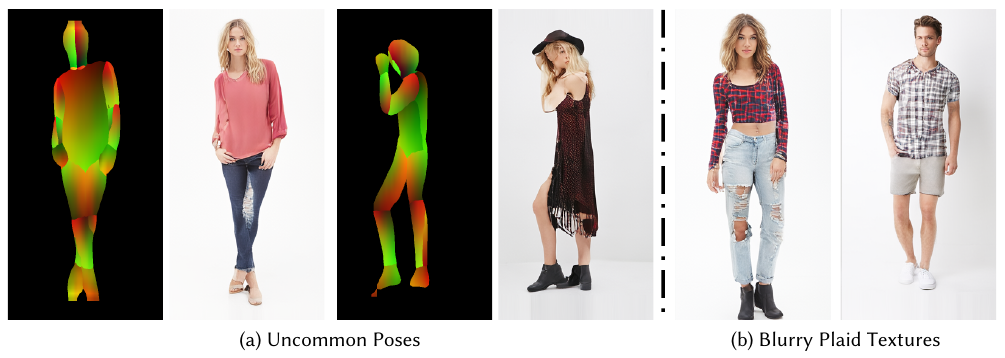}
  \end{center}
  \caption{\textbf{Failure Cases.} (a) Under uncommon poses, implausible artifacts will appear in the generated images. (b) Blurry plaid textures are generated due to imbalanced datasets.}
  \label{fig:failure}
\end{figure}
1) \textbf{Uncommon poses}. The performance would degrade with human poses which are uncommon in the DeepFashion-MultiModal dataset. Two examples of uncommon poses are shown in Fig.~\ref{fig:failure}(a). The first pose is with two legs crossed, artifacts would appear in the cross-region. The second person stands facing the side rather than the front. In this case, artifacts would appear in the face region, as the model is prone to generate faces heading the front.
And thus, the generated image looks unnatural.
Our framework is data-driven and can benefit from more diverse human datasets in future work.
2) \textbf{Plaid textures} are blurry as shown in as shown in Fig.~\ref{fig:failure}(b). This is attributed to the imbalanced textures in DeepFashion. Only 162 out of 10335 training images have plaid patterns in upper clothes. This is a common problem for all baselines, and our performance is superior. In future work, the performance could be boosted by adding more data with such complicated patterns. For newly added data, the labels for clothes attributes could be provided by the attribute predictor trained on our dataset. Some techniques dealing with imbalanced data could also be employed to mitigate the problem.
3) \textbf{Potential error in word embeddings.} Translating text descriptions to one-hot embeddings inevitably introduces errors. For example, for the length of sleeves, we only define four classes, \ie, sleeveless, short sleeves, medium sleeves, and long sleeves. If the user wants to generate a sweater with sleeves covering the elbow but not reaching the wrist, the synthesized human parsing cannot be perfectly aligned with the text inputs as the predefined texts cannot handle sleeves with arbitrary lengths.
In future work, continuous word embeddings could be employed to provide richer and more robust information.
\section{Conclusions}

In this work, we proposed the Text2Human framework for text-driven controllable human generation in two stages: pose-to-parsing and parsing-to-human. The first stage synthesizes the human parsing masks based on required clothes shapes. In the second stage, we propose a hierarchical VQVAE with texture-aware codebooks to capture the rich multi-scale representations for diverse clothes textures, and then propose a sampler with mixture-of-experts to sample desired human images conditioned on the texts describing the textures. To speed up the sampling process of hierarchical VQVAE and further refine the sampled images from the coarse level, a feed-forward codebook index prediction network is employed. Our proposed Text2Human is able to generate human images with high diversity and fidelity in clothes textures and shapes. We also contribute a large-scale dataset, named DeepFashion-MultiModal dataset, for the controllable human image generation task.

\begin{acks}
The authors would like to thank Zhuowei Chen, Kelvin C.K. Chan and Ziqi Huang for discussions and proofreading.
This study is supported by NTU NAP, MOE AcRF Tier 1 (2021-T1-001-088), and under the RIE2020 Industry Alignment Fund – Industry Collaboration Projects (IAF-ICP) Funding Initiative, as well as cash and in-kind contribution from the industry partner(s).
\end{acks}

\bibliographystyle{ACM-Reference-Format}
\bibliography{sample-bibliography}

\end{document}